\pdfoutput=1

\documentclass[11pt]{article}

\usepackage{acl}

\usepackage{times}
\usepackage{latexsym}
\usepackage[export]{adjustbox}

\usepackage[T1]{fontenc}

\usepackage[utf8]{inputenc}

\definecolor{light-gray}{gray}{0.90}
\definecolor{dark-gray}{gray}{0.30}
\definecolor{ourlightblue}{HTML}{E0ECF7}
\definecolor{ourdarkblue}{HTML}{092E6B}
\definecolor{msgrblue}{HTML}{4889f4}
\definecolor{msgrgray}{HTML}{e1e1e7}
\definecolor{msgrpaleblue}{HTML}{a9c6f5}
\definecolor{palegreen}{HTML}{c0eeC3}
\definecolor{palepurple}{HTML}{e5d1f8}
\definecolor{paleorange}{HTML}{f9dbb1}
\definecolor{palered}{HTML}{f8aca7}
\definecolor{paleyellow}{HTML}{f8f1a7}
\definecolor{brightergreen}{HTML}{63ec99}

\definecolor{botc}{rgb}{0.458, 0.488, 0.978}

\definecolor{emilycolor}{HTML}{00D0A1}
\definecolor{jxcolor}{HTML}{b94e48}
\def\ED#1{{\color{emilycolor}ED: \it #1}}

\usepackage{microtype}

\usepackage{algorithm}
\usepackage{algorithmic}
\usepackage{microtype}
\usepackage{amsmath}
\usepackage{multirow}
\usepackage{amsmath,array,graphicx,amssymb}
\usepackage[export]{adjustbox}
\usepackage{subcaption}
\usepackage{multirow}
\usepackage{mathtools} 
\usepackage{tabularx}
\usepackage{breakcites}
\usepackage{xcolor}
\usepackage{array}
\usepackage{makecell}
\usepackage{tabularx}
\usepackage{mathrsfs}
\usepackage{scalerel}
\usepackage{soul}
\usepackage{colortbl}
\definecolor{mygray}{gray}{.95}
\usepackage{soul}
\usepackage{booktabs}

\usepackage{graphicx}
\usepackage{caption}
\usepackage{algorithm}
\usepackage{algorithmic}
\usepackage{microtype}
\usepackage{amsmath}
\usepackage{multirow}
\usepackage{amsmath,array,graphicx,amssymb}
\usepackage{caption}

\usepackage{amsthm}
\usepackage{xspace}

\newcommand\blfootnote[1]{%
  \begingroup
  \renewcommand\thefootnote{}\footnote{#1}%
  \addtocounter{footnote}{-1}%
  \endgroup
}
\def \pipelinename{\textsc{Juicer}\xspace}
\newcommand{\wyshi}[1]{\textcolor{blue}{[\textbf{wyshi}: #1]}}
\def\TODO#1{{\color{red}TO DO: \it #1}}

\usepackage{scalerel,xparse}

\NewDocumentCommand\lemontitle{}{
    \includegraphics[scale=0.06]{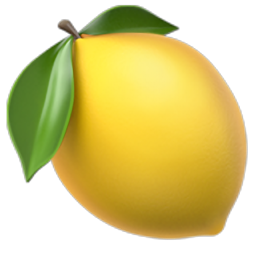}
}
\NewDocumentCommand\cherrytitle{}{
    \includegraphics[scale=0.06]{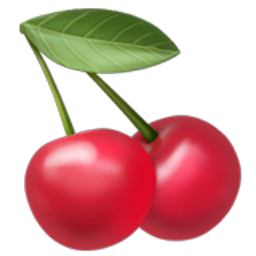}
}

\NewDocumentCommand\lemon{}{
    \includegraphics[scale=0.05]{EmojiFolder/lemon.png}
}
\NewDocumentCommand\up{}{
    \includegraphics[scale=0.7]{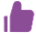}
}
\NewDocumentCommand\down{}{
    \includegraphics[scale=0.7]{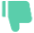}
}
\NewDocumentCommand\cherry{}{
    \includegraphics[scale=0.05]{EmojiFolder/cherry.png}
}
\NewDocumentCommand\question{}{
    \includegraphics[scale=0.05]{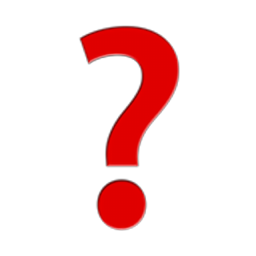}
}
\usepackage{adjustbox}
\usepackage{graphicx}
\usepackage{multirow}
\usepackage{booktabs}
\usepackage{todonotes}
\usepackage{float}
\usepackage{xstring}
\usepackage{enumitem}
\usepackage[bottom]{footmisc}

\definecolor{lightblue}{HTML}{E0ECF7}
\definecolor{darkblue}{HTML}{092E6B}

\title{When Life Gives You Lemons \lemontitle, Make Cherryade \cherrytitle: \\\textit{Converting Feedback from Bad Responses into Good Labels}
}
\author{Weiyan Shi$^{\dagger}$\\
  Meta AI \& Columbia University\\
  \And
  Emily Dinan\\
  Meta AI\\
  \And
  Kurt Shuster\\
 Meta AI \\
  \AND
  Jason Weston$^*$\\
  Meta AI \\
  \And
  Jing Xu$^*$ \\
  Meta AI \\
}

\begin{document}
\maketitle
\begin{abstract}
Deployed dialogue agents have the potential to integrate human feedback to continuously improve themselves.  However, humans may not always provide explicit signals when the chatbot makes mistakes during interactions. 
In this work, we propose  \pipelinename, a framework to 
make use of both binary and free-form textual human feedback.
It works by: (i) extending sparse binary feedback by training a \textit{satisfaction classifier} to label the unlabeled data; and (ii) training a \textit{reply corrector} to map the bad replies to good ones.
%
%
We find that augmenting training with model-corrected replies improves the final dialogue model, and we can further improve performance by using both  positive and negative replies through the recently proposed \textsc{Director} model. \blfootnote{$\dagger$ Work done when interning at Meta AI.}
\blfootnote{* Equal contribution.}
\end{abstract}







\section{Introduction}
Existing dialogue models are primarily trained on human-human conversations \cite{conneau2019unsupervised,baumgartner2020pushshift,smith2020bst}. 
As dialogue agents become increasingly powerful and carry substantial conversations with humans \cite{shuster2022blenderbot}, it becomes pressing to have the models learn from dialogue successes and failures  in the wild, 
and hence 
improve after deployment. 

Prior work has studied how to collect and learn from feedback in human-model dialogues \cite{li2016dialogue,li2016learning, hancock2019learning, xu2022learning}. But most existing methods were proposed under  settings where  either feedback can be obtained whenever needed or all turns 
are annotated with human feedback. For instance, \citet{xu2022learning} introduced a dataset with all turns annotated by crowdworkers with three types of feedback: (1) binary thumbs up/down; (2) free-form textual feedback on what went wrong; (3) gold corrections on what the bot should have said instead.   
Unfortunately, 
annotations such as thumb ups/downs and gold corrections 
are often sparse in real-life deployment settings.
For example, human conversationalists give thumbs up/down to bot messages in conversations with the deployed BlenderBot3 model around 5-6\% of the time  \cite{shuster2022blenderbot}.
On the other hand, human conversationalists may express their dissatisfaction with bad responses and explain what went wrong more naturally in free-form textual feedback as part of the conversation, rather than providing the exact gold corrections to those bot responses. 
Therefore, in this paper we study how to utilize sparse binary and gold correction feedback, and relatively dense free-form textual feedback to improve dialogue models during deployment.

In this work, we introduce \pipelinename, a framework to ``squeeze the juice'' out of the sparse human feedback in human-model conversations to improve the dialogue models after deployment. 
\pipelinename consists of four steps: (1) we first train a binary \textit{satisfaction classifier} and a \textit{reply corrector} on existing binary feedback and gold corrections; (2) we then use the \textit{satisfaction classifier} to label all the bot responses that are missing human labels;
(3) next we use the \textit{reply corrector} to correct  bad bot responses (lemons \lemon)  into good ones,  conditioning on human textual feedback; (4) finally we augment the training data with  the new good responses (cherryade \cherry)
and re-train our final dialogue models.

To evaluate \pipelinename on state-of-the-art chatbots in such a setting,
we thus construct a new sparse sampled version of the existing FITS dataset from \citet{xu2022learning}, which consists of fully annotated human-model conversations between users and existing state-of-the-art internet-augmented models such as BlenderBot 2 \cite{komeili2021internet,xu2021beyond} and SeeKeR \cite{shuster2022language}. 

We explore a variety of methods to take advantage of limited human feedback at each step of the \pipelinename framework. Our 
main results are:
\begin{itemize}
     \item We show that free-form textual feedback is a very useful signal for improving the performance of both a \textit{satisfaction classifier} to identify good and bad responses, and a \textit{reply corrector} to generate better corrections.
    \item Augmenting training data with \textit{reply-corrector}-generated corrections 
    works better than only training with existing gold corrections.
    \item Models such as \textsc{Director} \cite{arora2022director} that utilize both gold/predicted good and bad responses further improves the final dialogue model.  Our final best models outperform the baseline BlenderBot 2 model or using \textsc{Director} alone.
\end{itemize}


\section{Related Work}
Many recent works have studied how to align language models with human feedback  \cite{nakano2021webgpt,ouyang2022training,scheurertraining2022,saunders2022self, schick2022peer}. 
For instance,  InstructGPT \cite{ouyang2022training}  was fine-tuned using feedback from  labelers who ranked model outputs. 
\citet{scheurertraining2022} fine-tuned GPT-3 and InstructGPT on 100 examples of free-form textual feedback from humans to improve summarization tasks and found that only the larger models such as GPT-3 (175B) \cite{brown2020language} can generate accurate refinements using feedback.  \citet{saunders2022self}
fine-tuned large language models to generate self-critiques for summarization tasks to assist human annotators, and continued to refine the models on feedback.  In this work, we focus on improving dialogue agents given various human feedback signals (binary,  free-form natural language and gold corrections) and compare our methods to
\citet{scheurertraining2022}.


Existing works have also studied how to correct language model output.
For instance, \citet{elgohary2021nl} proposed a model to understand natural language feedback and produce a series of edits to correct a text-to-SQL semantic parser.  \citet{tandon2022learning}  trained a  memory-augmented corrector to convert feedback to edits and fix model outputs for a script generation task. 
 Some recent large language model research can also repair  generations given human feedback \cite{scheurertraining2022,saunders2022self}. 

Past research has also  explored how to integrate feedback into dialogue agents \cite{li2016dialogue,li2016learning, hancock2019learning,shuster2020deploying, xu2022learning}. \citet{li2016dialogue} investigated how to improve the chatbot's question-answering ability with general textual feedback in a reinforcement learning setting. \citet{hancock2019learning} developed a self-feeding chatbot that can construct new examples from existing human-bot conversations and ask for feedback when necessary to improve itself. \citet{xu2022learning} proposed a dataset with internet-augmented dialogues, where each turn is annotated with  human feedback, and they found that continuously retraining the model on binary feedback after deployment is helpful.  Our work 
focuses on converting  bad responses into good ones to augment the data and learn from feedback about failures.

\begin{figure*}[t!]
    \centering
    \includegraphics[width=1.0\textwidth]{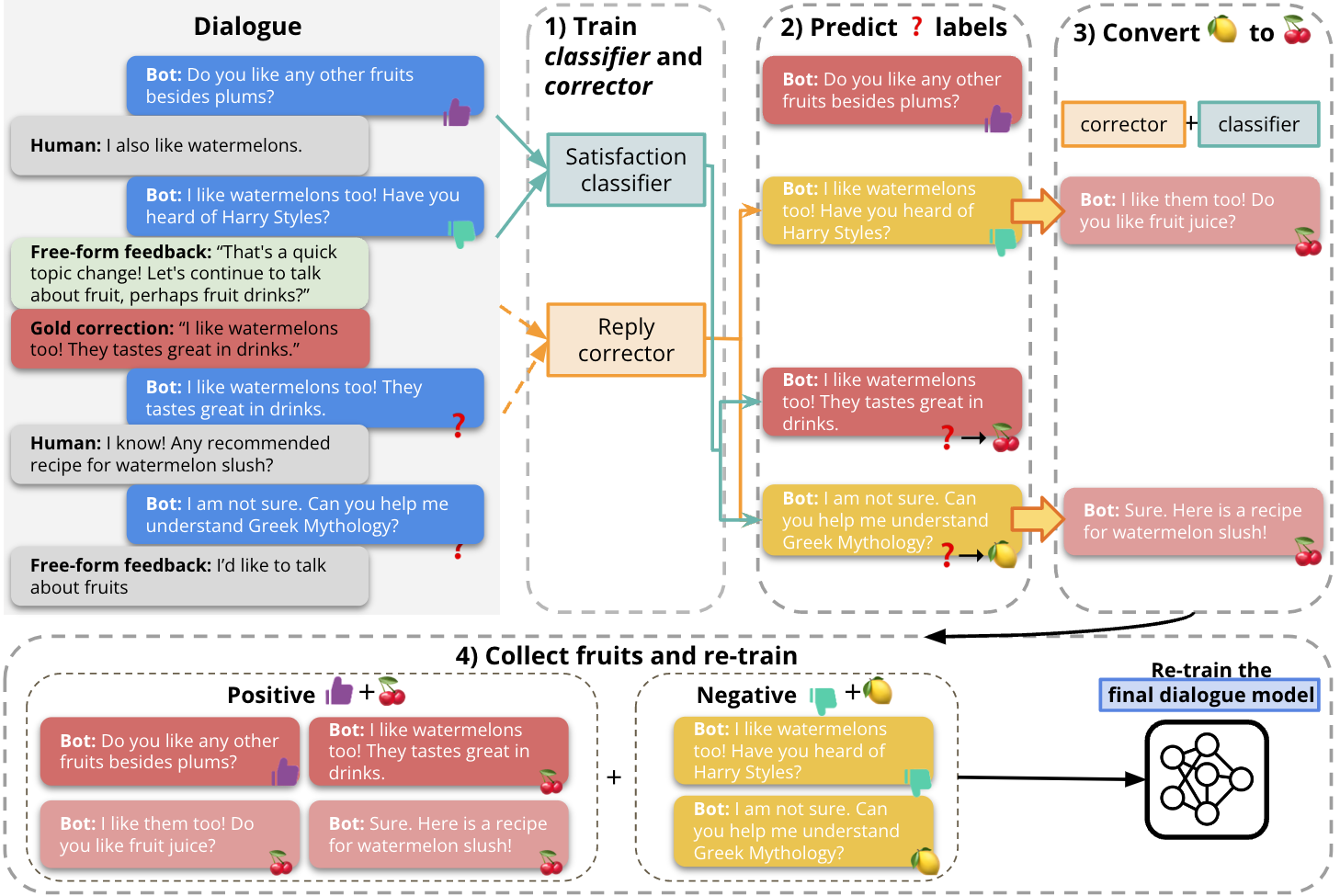}
    \caption{Our \pipelinename model. During deployment, we collect three types of human feedback: (1) binary thumbs up and down; (2) free-form textual feedback on what was wrong with the reply (\textit{``That's a quick topic change! Let's continue to talk about fruit, perhaps fruit drinks?''}); (3) gold corrections of poor replies (\textit{``I like watermelons too! They tastes great in drinks.''}). In \pipelinename, (1) we first train a \textit{satisfaction classifier} and a \textit{reply corrector} on existing feedback, (2) we then use the \textit{satisfaction classifier} to predict binary satisfaction labels for the un-annotated turns, (3) next we use the \textit{reply corrector} to convert the bad replies to good ones, (4) finally we collect the good and bad replies including corrections and re-train the final dialogue model to improve it with human feedback.   
    }
    \label{fig:main}
\end{figure*}

\section{Human Feedback Setting}
 
 As illustrated in the dialogue example in Figure~\ref{fig:main}, 
 we consider a deployed system where one can  collect three types of feedback: 
 \begin{itemize}
\item[(1)] \textbf{binary feedback}, where the human conversationalist explicitly likes (\up)  or dislikes (\down) a bot response; 
\item[(2)] \textbf{free-form textual feedback}, where the human  explains conversationally what was wrong when they dislike a response (e.g., \textit{``That's a quick topic change! Let's continue to talk about fruit, perhaps fruit drinks?''}); 
\item[(3)] \textbf{gold correction}, where the human conversationalist suggests an alternative reply the bot should have said, 
 (e.g., \textit{``I like watermelons too! They tastes great in drinks.''}). 
 \end{itemize}
 In a deployment setting, it is unnatural to ask users to always click the thumbs up and down and provide gold corrections whenever the bot makes a mistake. Instead, users tend to provide free-form textual feedback on what was wrong in their dialogue response to express dissatisfaction when the bot makes errors \cite{see2021understanding}.  Therefore many responses may be  missing binary feedback \cite{shuster2022blenderbot}. 
In this paper, we consider a sparse 
thumbs up/down signal and sparse 
gold correction signal setting,  but a dense
free-form textual feedback signal (i.e., mistakes are followed by textual feedback).
  After collecting conversations with these feedback signals, we can consider methods to utilize them  to improve the dialogue model.

\section{The {\sc Juicer} Method}

Figure~\ref{fig:main} shows the overview of our framework \pipelinename  to incorporate limited human feedback to improve the deployed dialogue model. The framework consists of training a \textit{satisfaction classifier}, a \textit{reply corrector}, and eventually the final dialogue model itself.
We define the notation here. For a given bot reply:
(1)\question denotes un-annotated turns;  (2)\up and \down: annotated as good or bad responses by users, as defined before; (3)\lemon: predicted as bad by the \textit{satisfaction classifier}; (4)\cherry: predicted as good by the \textit{satisfaction classifier}.

\pipelinename involves four steps, summarized here:

\begin{enumerate}
\item \emph{\textbf{Step 1.}} 
Train two supervised models: a \textit{satisfaction classifier} to detect good and bad replies, and a \textit{reply corrector} to correct the bad replies to good ones. 
\item \emph{\textbf{Step 2.}} Apply the \textit{satisfaction classifier} to predict binary labels for all replies missing binary feedback 
(\question$\rightarrow$\cherry or \lemon). After this step, each bot reply has a label.  
\item \emph{\textbf{Step 3.}} Use the \textit{reply corrector} to convert  the bad replies (those that are either disliked by human users or are predicted as bad by the \textit{satisfaction classifier} in \textit{Step 2} to good replies (\down$\rightarrow$\cherry, \lemon$\rightarrow$\cherry). 
\item \emph{\textbf{Step 4.}} Re-train the final dialogue model by augmenting the training data with 
the good (\up+\cherry) and bad  (\down+\lemon) replies derived from human feedback and the predictions from the previous steps.  
\end{enumerate}
Now we describe each step in more detail.


\subsection{Step 1: Train \textit{satisfaction classifier} and \textit{reply corrector} on existing feedback}

We first train two models: (1) a \textit{satisfaction classifier}, and (2) a \textit{reply corrector} in order to build an augmented training set in later steps. In our experiments, both models are trained with human-labeled data which come from the FITS task \cite{xu2022learning}, described further in  Section~\ref{sec:fits}.

\paragraph{(1a) \textit{Satisfaction classifier}}  The training target of the \textit{satisfaction classifier} is a binary satisfaction label (\up or \down). For the input to the classifier, we experimented with two variants: (1) the context + the bot reply to be labeled, and (2) the context + the bot reply to be labeled + the next human response. As shown in the example in Figure~\ref{fig:main}, when the first bot reply is given a thumbs-up, 
the next human response is a natural continuation of the conversation (e.g., \textit{``I also like watermelons''}); when the bot reply is disliked (the second bot reply), the next human response is free-form textual feedback on what went wrong (e.g., \textit{``That's a quick topic change! Let's continue to talk about fruit, perhaps fruit drinks?''}). Hence, the next human response can be indicative of the quality of its preceding bot reply, and we include it in the input. In our experiments, the \textit{satisfaction classifier} is trained by fine-tuning a 311M-parameter transformer pre-trained on pushshift.io Reddit data \cite{baumgartner2020pushshift}.

\paragraph{(1b) \textit{Reply corrector}} The input to the \textit{reply corrector} is the context + the bad bot reply to correct + the next human free-form textual feedback on what went wrong. The training target is the correction to the bad reply which can be either (1) gold corrections written by crowdworkers; or (2) the next bot replies from the original FITS data that are classified as good (``self-corrections''). We fine-tuned the \textit{reply corrector} from a 3B parameter R2C2 transformer model \cite{shuster2022language}. 

\subsection{Step 2: Predict missing labels }
In a bot-human dialogue, the binary feedback can be quite sparse,
with many replies having no explicit feedback.
We thus predict labels for these replies with the \textit{satisfaction classifier} trained in Step 1a.
After this step, every bot reply in the dataset has a binary label either from the original human binary feedback (\up or \down), or predicted by the \textit{satisfaction classifier} (\cherry or \lemon).  

\subsection{Step 3: Convert lemons to cherries}

We can now augment the training data. We  use the \textit{reply corrector} trained in step 1b to generate improved replies for any examples labeled as bad (\down or \lemon), and then add them to the training set for the final dialogue model.

\paragraph{Selecting correctable cases} However, we note that not all bad responses are easily correctable given free-form textual feedback. For example, the human feedback ``You are talking nonsense!'' could help indicate this is a \lemon using the \textit{satisfaction classifier}, but is less helpful for knowing what the right response is, compared to more constructive feedback such as ``That's a quick topic change! Let's continue to talk about fruit, perhaps fruit drinks?''
We thus experiment with detecting cases that are ``correctable'', and only use these to augment our training data.
We first embed the free-form textual feedback and the immediate next bot reply in recorded conversations 
with Sentence-BERT \cite{reimers2019sentence}, and then calculate their cosine similarity score. If the score is high, it means 
that the human free-form textual feedback is easier for a model to comprehend and thus revise its own response accordingly. We define such examples as correctable and then threshold the similarity score to pick out 
correctable cases. 

\paragraph{Predicting reply corrections}
To obtain the corrections, we adopt a reranking-based learning method widely used in many previous studies \cite{nie2020like,nakano2021webgpt, askell2021general} to score and rank the generations. We first use the \textit{reply corrector} to generate many correction candidates (60 in our experiments). Then we concatenate the original context with the correction candidates  and feed them into the \textit{satisfaction classifier} from Step 1a. 
Finally, we select the top one with the highest probability output by the classifier as the final correction. If all generated corrections are predicted as bad, we will skip this example.

\subsection{Step 4: Collect fruits and re-train}
After the previous steps, each bot response is annotated with either a gold or predicted binary label, and those labeled as bad are converted from bad responses to good ones using human feedback.  The final step is to 
augment the training set of the final dialogue model with the new data.

One straightforward method to improve the model is to augment the training data with  all the positive replies including the corrections (\up + \cherry) and use the standard language modeling objective. However, this standard  training does not utilize negative/bad replies (\down + \lemon) to avoid them. We hence also apply the recently proposed {\sc Director} model \citep{arora2022director}  to both reinforce the positive responses and penalize the negative ones. {\sc Director} is a unified decoder-classifier model jointly trained with a language modeling task and a classification task. During inference, it uses its language modeling head to predict the next token probability, and its classifier head to decide if the tokens belong to positive examples to generate the final output. But it is worth noting that in this step in \pipelinename, we could use any other approach that utilizes both positive and negative responses to re-train and improve the final dialogue model.

\section{Experimental Setup}
In our experiments, we used the 3B parameter BlenderBot2 (BB2 3B) \cite{komeili2021internet, xu2021beyond} as the base dialogue model and try to improve it with human feedback from deployment.   

\subsection{Datasets: FITS and DEMO}

We performed experiments on the FITS \cite{xu2022learning} 
dataset.
We also tested the zero-shot transferability of both the \textit{satisfaction classifiers} and the \textit{reply correctors} on a real deployment dataset DEMO \cite{ju2022trolls}.

\subsubsection{FITS} \label{sec:fits}
FITS contains internet-augmented human-bot dialogues with annotated feedback for every turn, including a binary label, free-form textual feedback and a gold response, with around 39k bot utterances in total.  See Section~\ref{app:FITS} for more details.  
%
To mimic a deployment setting with limited feedback,  we uniformly sampled 20\% of the bot responses 
from the training set of FITS and considered them to have binary feedback and gold labels, while the rest were considered unlabeled. However, we did not remove free-form textual feedback when it is present, as it remains part of the conversation, see \autoref{fig:main}.
Table~\ref{tab:data stat} in the Appendix shows the data statistics after sampling. 

We used  the original FITS  validation, test set  and unseen test set (of new conversational topics) for evaluation, and employed the same metrics as \citet{xu2022learning} for the final dialogue models: perplexity, F1 overlap with the gold annotation, and human evaluation via conversations with the bot. During conversations, crowdworkers click \up or \down per turn and give a final rating (a score out of 5) in the end. We report  the average good  response rate in percentage. 

\subsubsection{DEMO}
The  dataset DEMO is from the deployment of BlenderBot 3 \cite{shuster2022blenderbot} with responses verified by  crowdworkers \cite{ju2022trolls}. 
In total  923 bot responses across 81 conversations are used as an evaluation set.

\subsection{Baselines}

We have two categories of baselines: (1) without model-augmented data, and (2) with a prompt-based \textit{reply corrector}. In addition, we also compare with oracle methods using 100\% labeled feedback data without sampling. 

\paragraph{Baselines without augmentation. 
} 

The most straightforward baselines are to fine-tune with the limited human-labeled feedback only.

\begin{itemize}
        \item \textbf{Gold corrections from 20\%} Gold corrections provide a strong learning signal. Here, we simply fine-tune BB2 3B on the 20\% gold corrections. 
        
    \item \textbf{Free-form textual feedback from 20\%}  Following \citet{hancock2019learning}, we fine-tune BB2 3B with the context as the input and the free-form textual feedback (identified as the response following the  bad \down responses)  as the target.  
\end{itemize}

\paragraph{Baseline with a prompt-based \textit{reply corrector}.} 

Instead of training a supervised \textit{reply corrector} with gold corrections, this baseline prompts an off-the-shelf model with free-form textual feedback and instructions like ``given the feedback, correct the original response'' as a \textit{reply corrector} to  generate corrections, and then fine-tunes  the final dialogue model on these corrections. 
\begin{itemize}
    \item \textbf{3B-all-corrections}: 
    \citet{scheurertraining2022} proposed an approach to improve language models with language feedback, originally applied to summarization tasks, which we adapt here for dialogue.  
    Given a small number (100) human feedback samples, they prompted a language model to condition on the context (input+feedback) to re-generate multiple summarization corrections, picked the correction with the highest similarity score with the feedback, and finally fine-tuned the language model on the corrections to improve it. In our implementation, we use the baseline BlenderBot 2 model (3B) as the \textit{reply corrector}. While \citet{scheurertraining2022} used larger language models (175B), our implementation of the baseline is  more comparable to our \pipelinename models since  our \textit{reply corrector} also has 3B parameters.  
    In our experiments, instead of using only 100 examples, we make this a stronger baseline by generating corrections for \textit{all} the bad replies. 
\end{itemize}
\vspace{-1em}

\begin{table*}[!htb]
\centering
    \begin{subtable}{.975\textwidth}
      \centering
\begin{adjustbox}{width=.9\textwidth}

\begin{tabular}{l|cc|cc|cc|cc}
\toprule
\midrule
\textbf{\underline{(1a) \textit{Satisfaction Classifier}}}   & \multicolumn{2}{c|}{\textbf{Valid}} &
                       \multicolumn{2}{c|}{\textbf{Test}}&
                       \multicolumn{2}{c|}{\textbf{Test Unseen}}&
                       \multicolumn{2}{c}{\textbf{DEMO (zero-shot)}}
                       \\
        \textbf{Input}               & \textbf{Acc$\uparrow$}        &  \textbf{F1$\uparrow$}     
        & \textbf{Acc$\uparrow$}                      &  \textbf{F1$\uparrow$} & \textbf{Acc$\uparrow$}                       &  \textbf{F1$\uparrow$} & \textbf{Acc$\uparrow$} & \textbf{F1$\uparrow$} \\
                       \midrule
                       \midrule
   context+bot+human &\textbf{94.66}&	\textbf{97.25}&	
\textbf{95.76}&	\textbf{97.83}&	\textbf{96.74}&	\textbf{98.34}&	\textbf{59.73}&	\textbf{71.24}\\
   context+bot & 75.58&	86.07&	
   74.53&	85.38&	71.46&	83.25&	56.60&	64.77\\ 
\bottomrule
                                              
\end{tabular}
\end{adjustbox}
\caption{\textit{Satisfaction classifier} results (classification balanced accuracy and balanced f1) on both FITS and DEMO (zero-shot). Adding the next human message helps the satisfaction prediction, even in the zero-shot case.  
}
\label{tab:classifier}


    \end{subtable}%
    \qquad
    \qquad
    \newline
    
    \begin{subtable}{.95\textwidth}
      \centering
      \begin{adjustbox}{width=.9\textwidth}
\begin{tabular}{l|l|l||l|l|l|l}
\toprule
\midrule
\textbf{\underline{(1b) \textit{Reply Corrector}}}&  \multicolumn{2}{c||}{\textbf{Valid}} & \multicolumn{2}{c|}{\textbf{Test}}                                & \multicolumn{2}{c}{\textbf{Test Unseen}} \\
\textbf{Input}& \multicolumn{1}{c}{\textbf{F1$\uparrow$}}    & \multicolumn{1}{c||}{\textbf{PPL$\downarrow$}}      & \multicolumn{1}{c}{\textbf{F1$\uparrow$}}    & \multicolumn{1}{c|}{\textbf{PPL$\downarrow$}}  & \multicolumn{1}{c}{\textbf{F1$\uparrow$}} & \multicolumn{1}{c}{\textbf{PPL$\downarrow$}} \\
\midrule
\midrule
\multicolumn{7}{l}{\textbf{Oracle with 100\% annotations}} \\
\midrule
gold corrections from 100\% 
&    23.39&	2.93&		21.83&	2.63&	22.27&	4.56                             \\
\midrule
\multicolumn{7}{l}{\textbf{w/ free-form textual feedback}} \\
\midrule
 gold corrections from 20\%  + self-corrections
&  \textbf{21.41}                  & \textbf{3.07} & \textbf{20.20} &	\textbf{2.75}&	\textbf{21.77}&	\textbf{4.66}                 \\

 gold corrections from 20\%                                                                  &  17.10      & 3.37  & 16.21&	2.98&	17.91&	4.97      \\
\midrule
\multicolumn{7}{l}{\textbf{w/o free-form textual feedback}} \\
\midrule

 gold corrections from 20\% + self-corrections                                                                  &  18.80      & 3.13  &18.36 & 2.82	&18.97&4.84	    \\

 gold corrections from 20\%                                                                  &  16.41      & 3.40  &15.08 &3.04&16.46	&5.06	      \\

\bottomrule
\end{tabular}
\end{adjustbox}
\caption{\textit{Reply corrector} perplexity and F1 on valid/test/test unseen sets.
Augmenting with self-corrections improves the result,  comparable to the oracle model using 100\% gold corrections.  Using free-form textual feedback is helpful. 
}
\label{tab:reply corrector}

    \end{subtable}%

    \caption{Performance of the modules in Step 1: (a) \textit{satisfaction classifier}, and (b) \textit{reply corrector}.  }
\label{tab:satisfaction classifier + reply corrector}
\vspace{-1em}
\end{table*}

\subsection{\pipelinename models} 
We also compare several variants of \pipelinename. 
\begin{itemize}
    \item \textbf{\pipelinename}. We fine-tune BB2 3B by augmenting the 20\% human-annotated data with (1) predicted good responses by the \textit{satisfaction classifier} from the remaining 80\% un-annotated turns, and (2) predicted corrections generated by the \textit{reply corrector}, filtered to only include the correctable cases rather than using all the predicted corrections.     
    
    \item \textbf{\pipelinename} + \textbf{\textsc{Director}}. We fine-tune using {\sc Director} which uses both the positive and negative replies.  
    Both gold annotations and the filtered corrections generated by the \textit{reply corrector} are used as positive classification data. Bad responses labeled by humans or the \textit{satisfaction classifier} are used as negative data for fine-tuning the classifier head. 

    \item \textbf{w/o predicted corrections (from Step 3)}. In this ablation, we fine-tune the final dialogue model with only predicted good responses by the \textit{satisfaction classifier}, without the  corrections generated by the \textit{reply corrector}. 
    \item \textbf{w/o selecting correctable cases}. 
    In this ablation, we only augment  with (1) predicted good responses by the \textit{satisfaction classifier}, and (2) all the predicted corrections without selecting the more correctable cases. 
    This tests if selecting correctable cases brings improvements.

\end{itemize}

\section{Results}

We first evaluate the \textit{satisfaction classifier} (Table~\ref{tab:classifier}), and the \textit{reply corrector} (Table~\ref{tab:reply corrector}). We then perform both automatic and human evaluations on the final dialogue models (\autoref{tab:dialogue model_auto_eval} and \autoref{tab:dialogue model_human_eval}).

\subsection{\textit{Satisfaction classifier}}
\label{sec:satisfaction classifier}
Table~\ref{tab:classifier} shows the classifiers' performance on the FITS data and also their zero-shot performance on DEMO.

\paragraph{Adding the next human response helps.} 
We find the balanced accuracy of detecting
satisfaction using only the dialogue context and the bot response itself is $\sim$75\% on FITS.
It is significantly improved to $\sim$95\% by including the next human message in the input. A similar improvement is found when measuring balanced F1 as well. 
On the deployment dataset DEMO where organic users are not required to always write free-form textual feedback when seeing a bad reply, adding the human response still improves the balanced F1 from 64.77 to 71.24, despite this being zero-shot performance (without training on this dataset). These results indicate the importance of using the next human message to make satisfaction classification decisions.
As using the next human response helps, we default to using this \textit{satisfaction classifier} variant in our standard
\pipelinename setup.

\subsection{\textit{Reply corrector}}
Table~\ref{tab:reply corrector} shows the results of training the \textit{reply corrector}, comparing different input feature choices.

\paragraph{Free-form textual feedback improves the correction.} We performed an ablation study where the \textit{reply corrector} trains on  (context + bad reply $\rightarrow$ \textit{good reply}) without the free-form textual feedback, shown in ``w/o free-form textual feedback''. As expected, adding free-form textual feedback on what went wrong improves the \textit{reply corrector}'s performance. The best results are relatively close to the oracle performance which uses 100\% (rather than 20\%) gold data for training (23.39 F1 vs. 21.41 and 2.93 PPL vs. 3.07).

\paragraph{Augmenting with self-correction pairs helps.} 
The standard \textit{reply corrector}  trains on ``gold-correction'' pairs (context  + bad reply + free-form textual feedback$\rightarrow$ gold correction). Besides these human-written gold corrections, we can also train the \textit{reply corrector} on ``self-correction'' pairs (context + bad reply + free-form textual feedback$\rightarrow$ \textit{good bot reply}), where a bad reply is followed by a good bot reply either liked by humans or predicted  as good by the \textit{satisfaction classifier}, suggesting that  the bot ``corrects'' itself in the following turn.  We found that augmenting with these ``self-corrections'' improves the F1 from 17.10 to 21.41. We can also multitask with various dialogue tasks to further improve the \textit{reply corrector}'s performance. See Section~\ref{appendix:multitaking in reply corrector} for more details.  

\paragraph{Qualitative results show the corrections make sense.} We also include generated correction examples on the FITS data in Appendix Table~\ref{tab:qualitative on FITS} and on the deployment data in a zero-shot fashion in Appendix Table~\ref{tab:qualitative on deployment}. These examples show that the \textit{reply corrector} can integrate free-form textual feedback to correct the bad replies, even for zero-shot deployment data. 

See section~\ref{appendix:eval reply corrector} for further details and results on the {\em reply corrector} evaluation. 

\begin{table*}[t!]
\centering
\begin{adjustbox}{width=.7\textwidth}
\begin{tabular}{l|cc|cc|cc}
\toprule
\midrule
                                                       
 \multirow{1}{*}{\textbf{\underline{Final dialogue model}}}  & \multicolumn{6}{c}{\textbf{Automatic  evaluation}}    \\ 
 &\multicolumn{2}{c|}{\textbf{Valid}} & \multicolumn{2}{c|}{\textbf{Test}} &        \multicolumn{2}{c}{\textbf{Test Unseen}} \\
& \textbf{F1$\uparrow$} & \textbf{PPL$\downarrow$} & \textbf{F1$\uparrow$} & \textbf{PPL$\downarrow$} & \textbf{F1$\uparrow$} & \textbf{PPL$\downarrow$}   \\
\midrule
\midrule
\textcolor{black}{BB2 3B}   & 14.4 & 10.6 & 14.7 & 10.3 & 15.3 & 9.3 \\
\textcolor{black}{~~+gold corrections from 20\%} & 16.2  & 9.1  &15.6	&8.9&	17.9&	8.4  \\
 ~~+free-form textual feedback from 20\%      & 13.1  & 10.4 & 12.6&	10.3&	13.7&	9.6 \\
 \midrule
 3B-all-corrections (prompt-based)  & 14.2 & 8.9 & 14.5 &	8.7 &	15.2 &	8.2  \\
\midrule
\midrule
\multicolumn{5}{l}{\textbf{\pipelinename models}}  \\
\midrule
\midrule
~~+\pipelinename
& 16.7       & \textbf{8.5} & 16.2	&8.4 &	\textbf{18.5}	&\textbf{8.0} \\
~~+\pipelinename+{\sc Director} & \textbf{17.2} &{\small n/a}&\textbf{16.7}& {\small n/a}		&17.7 &{\small n/a}\\
\midrule
\midrule
\multicolumn{6}{l}{\textbf{\pipelinename ablations}}    \\
\midrule
\midrule
w/o predicted corrections         &  15.7       & 9.0   & 15.8 &	8.8 &	17.9 &	8.2    \\
w/o selecting correctable cases  & 16.4 & 8.5 & 16.4 &	8.4 &	18.0 &	8.1 \\
\bottomrule
\end{tabular}
\end{adjustbox}
\caption{Final dialogue model automatic evaluation results. 
All the dialogue models are fine-tuned from BB2 3B. \pipelinename models with augmentations are better than the baselines without augmentations. \pipelinename with a supervised \textit{reply corrector} also performs better than the baseline with a prompted-based \textit{reply corrector}. 
\textsc{Director} utilizing negative examples is effective. Using predicted corrections and selecting correctable cases are useful.  
}
\label{tab:dialogue model_auto_eval}
\end{table*}


\begin{table}[t!]
\centering
\begin{adjustbox}{width=.47\textwidth}
\begin{tabular}{l|cc}
\toprule
\midrule
{\bf \underline{Final dialogue model}} &   \multicolumn{2}{c}{\textbf{Human evaluation}}   \\                 & \textbf{Good\% $\uparrow$}  & \textbf{Rating $\uparrow$} \\
\midrule\midrule
\textcolor{black}{BB2 3B}    &33.2\% & 3.09 \\
\textcolor{black}{~~+gold corrections from 20\% }                                        & \textbf{39.4\%} &  2.89  \\

\midrule
\midrule
\multicolumn{3}{l}{\textbf{\pipelinename models}}                                      \\
\midrule
\midrule
~~+\pipelinename &   \textbf{41.9\%} & 3.06   \\
~~+\pipelinename+{\sc Director}  &   {\bf 45.5\%} & {\bf 3.34}  \\
\bottomrule
\end{tabular}
\end{adjustbox}
\caption{Final dialogue model human evaluation results.  We report the \% of good responses and the overall rating, as judged by crowdworkers during conversations.  We  bold statistically significant improvements (independent two-sample $t$-test, $p < 0.05$) of methods over the   BB2 3B baseline.
\pipelinename outperforms the baselines.  {\pipelinename+\sc Director} works the best.
}
\label{tab:dialogue model_human_eval}
\vspace{-1em}
\end{table}

\subsection{Final dialogue model evaluations}
The final dialogue model results are given in 
\autoref{tab:dialogue model_auto_eval} (automatic evaluations) and
\autoref{tab:dialogue model_human_eval} (human evaluations). 
%
All methods are fine-tuned from the 3B parameter BlenderBot 2 (BB2), making the models comparable. 


\paragraph{Using \pipelinename to augment data improves results.}
\pipelinename yields significant gains over the baseline transformer BB2 3B in both automatic evaluations
and human evaluations. For example, we see an F1 increase from 15.3 to 18.5 on the unseen test set, and an improvement of good responses from 33.2\% to 41.9\% in human evaluations.
\pipelinename  also performs better than baselines without augmentation (e.g., gold corrections from 20\%).
\vspace{-0.5em}

   
  \paragraph{Our supervised \textit{reply corrector} outperforms a prompt-based one.} 

 Compared to the prompt-based \textit{reply corrector} baseline \cite{scheurertraining2022}, all the \pipelinename models perform better in automatic evaluations. When the prompt-based model is used as a \textit{reply corrector} to produce corrections to augment the final dialogue model training, the final model evaluation (F1=14.2, ppl=8.9) is worse than augmenting with the corrections in \pipelinename (F1=16.7, ppl=8.5).

\paragraph{Augmenting training  with predicted corrections in \pipelinename helps.} 
\pipelinename augments training with predicted corrections, which  improves both the F1 and perplexity across the board compared to \pipelinename without predicted corrections, e.g. 18.5 vs. 17.9  on the test unseen F1. 
This makes sense because 
the  predicted corrections are generated by the  \textit{reply corrector} given human free-form textual feedback which contains valuable information, and fine-tuning the final dialogue model  on these corrections can steer it toward better replies. 


\paragraph{Selecting correctable cases can help.} 
\pipelinename picks only correctable cases to augment the training data, 
with around 62\% of cases selected (threshold chosen based on the validation set). 
Compared to naively augmenting with all predicted corrections, 
we see gains on valid and unseen test F1 (18.5 vs. 18.0), although there is no gain on the seen test set. 


\paragraph{\textsc{Director} provides further gains. } \textsc{Director} utilizes both the (predicted) binary feedback signal and textual feedback signal to avoid negative responses. Applying it improves the results further over standard \pipelinename (45.5\% good responses vs. 41.9\% for  \pipelinename~{\em without} {\sc Director}, as measured by human evaluations). Because \textsc{Director} uses a classifier head to decide if  a token should be included in the final generation, the distribution is altered and perplexity measures are not applicable. However, it  gives gains in F1 on valid and test sets, although not on the unseen test set. \pipelinename and \textsc{Director} together also outperforms  \textsc{Director} alone, even when  \textsc{Director} uses 100\% gold binary labels, see Appendix \autoref{tab: oracle performance}.
Further variants and experiments with   \textsc{Director} are  also given in Section~\ref{appendix:director overlap}.

\paragraph{\pipelinename achieves comparable results to methods with oracle access to gold labels. }
Compared to methods using 100\% gold data which was not given to \pipelinename, our best \pipelinename models  achieve comparable performance, especially on F1 and human evaluation. For example, test unseen F1=17.6 for the best ``oracle'' method vs. 17.7 for the best \pipelinename model, 47.0\% vs. 45.5\% good responses, and 3.38 vs. 3.34 in human ratings.  
See
Appendix Table \ref{tab: oracle performance} for further details. 
These ``100\% data'' methods can be seen as upper bound results, showing that \pipelinename does extract most of the signal from the portion of the dialogue data without binary or gold feedback.

See Section~\ref{appendix: final dialogue model} for 
further experiments and details on final model evaluations. 

\vspace{-0.5em}

\section{Conclusion}
Deployed dialogue agents should continuously improve by using human feedback gathered during interactions. Unfortunately, feedback collected in the wild can be limited.  In this paper, we proposed \pipelinename, a framework to efficiently use limited organic feedback signals (binary labels and gold corrections) if free-form textual feedback is provided.
\pipelinename works by correcting bad responses into good ones to augment the training data for the final dialogue model. 
Experiments show that augmenting with such predictions can integrate human feedback and improve overall performance. 

\section{Limitations and Discussions}
In our experimental setting, we assume dense free-form textual feedback, i.e., a bad reply is always followed by a free-form message explaining what was wrong. In real deployments, this free-form textual feedback signal may not always be given and without it, the binary \textit{satisfaction classifier} may not necessarily achieve a high accuracy or F1 (e.g., 90+), which could also impact the later steps. It remains to be seen  in real deployments 
how dense this signal is, and
what methods can be used to encourage users to make these signals as dense as possible, so that strong feedback signals are available to train on.

We have also assumed good intent from human conversationalists, but it is possible to have adversarial and bad actors interacting with the bot. In particular, incorrect feedback or opposite feedback (e.g., thumbs up instead of down) could be supplied by the human for incorrect bot behavior. We see this as an important research direction that should be pursued in parallel to work on algorithms like the ones we study here. See e.g. \citet{ju2022trolls} for recent work addressing bad actors and adversarial feedback.

The training/evaluation  loop of \pipelinename can be long due to its iterative nature. The advantage of using a \textit{reply corrector} is that we can qualitatively evaluate the quality of the generated corrections. But the drawback is that we need to first train a \textit{reply corrector}, use it to generate corrections, and finally improve the dialogue. We assume that the best \textit{reply corrector} will lead to the best final dialogue model, but this remains to be studied. Another possible direction is to use a latent \textit{reply corrector} to integrate the feedback in a more end-to-end fashion instead of a supervised \textit{reply corrector} that will generate explicit corrections separate from the dialogue model.  

Additionally, the proposed \pipelinename framework improves the dialogue model offline rather than correcting the response on the fly. With the necessary infrastructure support, there is potential for improving the models online. This could be a natural setting for reinforcement learning to get interactive feedback and iteratively update the model policy as the conversation continues.
Such a direction does not come without dangers, however, such as the model degrading if it receives poor inputs, e.g. from bad actors as mentioned before.

\bibliography{anthology,custom}
\bibliographystyle{acl_natbib}
\clearpage
\appendix
\section{Appendix}
\subsection{Details on FITS}
\label{app:FITS}
In this section, we describe the existing FITS task \cite{xu2022learning}
in more details. 
In FITS, each bot message is annotated with the following feedback. The setting also ensures that after the human provides the feedback, the conversation can be continued with the feedback integrated. 

\begin{itemize}
    \item A binary satisfaction label.
    \item If it is a bad reply, the human provides free-form textual feedback on what went wrong  in the next human message.
    \item Multiple-choice selection on what the bot could do to improve this turn:
        \begin{itemize}
        \item[(a)] using a better search query; or
        \item[(b)] attending better to the search results; or
        \item[(c)] other issues with the overall reply; or
        \item[(d)] no issue (a good reply).
    \end{itemize}
    \item If selecting (a), the human provides a better search query, which will be used in the next turn to continue the conversation.
    \item If selecting (b), the human is presented with the search results and selects the relevant sentences, which will be added to the model input in the next turn.
    \item If selecting (c), the human provides a better overall reply (a gold correction), which is copied to be the next bot response. 
\end{itemize}
\subsection{Sampling FITS}
In our experiments, we uniformly sample 20\% of the FITS training set to mimic a deployment setting with sparse binary and gold feedback.
Table~\ref{tab:data stat} shows the sampled dataset statistics. Out of the 20\% of the training set with labels, 1376 examples are ``better overall reply'' annotated with gold corrections, which accounts for 7\% of all bad responses to be corrected in FITS. Those 1376 corrections will later be used to train the \textit{reply corrector} for augmenting limited human feedback.
The size of 20\% of the FITS training set (7768 examples) is also similar to that of the validation and test sets. 

\begin{table}[]
\centering
\begin{adjustbox}{width=.98\columnwidth}
\begin{tabular}{l|c|c|c|c}
\toprule
\midrule
\textbf{Feedback Breakdown} & \textbf{Train (20\%)} & \textbf{Valid} & \textbf{Test} & \textbf{Test Unseen} \\
\midrule
\midrule
Total                & 7768  & 4245  & 9726 & 8907        \\
\midrule
Better Search Query  & 1056  & 605   & 1167 & 1036        \\
Better Results Usage & 1383  & 756   & 1527 & 1310        \\
Better Overall Reply & 1376  & 714   & 1493 & 1372        \\
Good Response        & 3953  & 2170  & 5539 & 5189   \\
\bottomrule
\end{tabular}
\end{adjustbox}
\caption{Data statistics of the sampled version of FITS used in our experiments. We sampled 20\% from FITS. Note that the training set size of labeled binary feedback is similar to the test sets.  }
\label{tab:data stat}
\end{table}

\subsubsection{Varying the sampling rate}
\label{app:vary sampling rate}

As an ablation study, we varied the sampling rate. 
Table~\ref{tab:varying sample rate} shows different final dialogue models' results with different sampling rates. The input to the dialogue model is the context, and the output is the human-written gold correction.

As we increase the sampling rate,  the final dialogue model's perplexity improves in general, but the gain becomes smaller. For instance, when we only sample 5\% from FITS, the validation perplexity is 9.52; if we increase the sampling rate to 20\%, the perplexity is 9.09; but when we further increase the sampling rate to 30\% and 50\%, the perplexity becomes  9.12 and 8.80 respectively. 

We find the 20\% sampling rate is a good balancing point with both reasonable F1 and perplexity.

\begin{table*}[]
\centering
\begin{adjustbox}{width=.6\textwidth}
\begin{tabular}{l|cc|cc|cc}
\toprule
\midrule
                    \textbf{\underline{Final dialogue model}}                                         & \multicolumn{2}{c}{\textbf{Valid}} & \multicolumn{2}{c}{\textbf{Test}} &        \multicolumn{2}{c}{\textbf{Test Unseen}} \\
                 Varying the sampling rate                         & \textbf{F1$\uparrow$} & \textbf{PPL$\downarrow$} & \textbf{F1$\uparrow$} & \textbf{PPL$\downarrow$} & \textbf{F1$\uparrow$} & \textbf{PPL$\downarrow$}                      \\
\midrule
\midrule
\multicolumn{7}{l}{\textbf{Baseline performance varying the sampling rate}}                                      \\
\midrule
~~+gold correction from 20\%                                          & 16.2  & 9.1  &15.6	&8.9&	17.9&	8.4   \\
\midrule
~~+gold correction from 5\% & 16.8 & 9.5 &16.6&9.3&18.9&8.5\\
~~+gold correction from 10\% & 16.6 & 9.3&16.5&9.1&18.9&8.4\\
~~+gold correction from 30\% & 15.7 & 9.1&16.3&8.6&18.1&8.3\\
~~+gold correction from 50\% & 16.5 & 8.8&16.0&9.0&17.9&8.3\\

\bottomrule
\end{tabular}
\end{adjustbox}
\caption{Final dialogue model results varying the sampling rate. Perplexities get better as we increase the sampling rate, but the gain becomes smaller. F1 first gets worse and then goes up. These suggest that a sampling rate of 20\% is a good balancing point with both a good F1 and a good perplexity. 
}
\label{tab:varying sample rate}
\end{table*}

\begin{table*}[]
\centering
\begin{adjustbox}{width=1.0\textwidth}
\begin{tabular}{l|m{6cm}|m{3cm}|m{3cm}|m{6cm}}
\toprule
\midrule
\textbf{Model}                   & \textbf{Inputs $\rightarrow$ Outputs}& \textbf{Fine-tuned from} &  \textbf{Evaluated on}  &                                               \textbf{Description}                                                        \\
\midrule
\midrule
(1a) \textit{Satisfaction classifier}&  Context + bot reply (+ the next human response) $\rightarrow$   binary \{good, bad\} on the bot reply & 311M Transformer & FITS valid\&test and DEMO 
& Given the context, a bot reply and potentially the next human message, detect if the bot reply is satisfactory \\
\midrule
(1b) \textit{Reply corrector}         & Context + bad bot reply + free-form textual feedback $\rightarrow$  improved reply (``a correction'') 
& 3B R2C2 \cite{shuster2022language} & Gold corrections in  FITS  valid\&test (on gold search results) &
Given the context, the bad reply, and free-form textual feedback, generate an improved reply (``correct the bad reply'')      \\
\midrule
(2) Final dialogue model    & Context $\rightarrow$  reply 
& 3B BlenderBot 2      &  Gold corrections in  FITS  valid\&test (on live search results) &
Given the context, generate a reply. We fine-tune our models using BlenderBot 2 as the base model \cite{komeili2021internet,xu2021beyond}.        \\     
\bottomrule                                                      
\end{tabular}
\end{adjustbox}
\caption{The input, output, and description of the three models used in \pipelinename. }
\label{tab:model description}\vspace{-1em}
\end{table*}

\subsection{Different modules in \pipelinename}
There are different modules involved in \pipelinename and we summarize them in Table~\ref{tab:model description}. To sum up, \pipelinename has two helper modules, a \textit{satisfaction classifier} and a \textit{reply corrector} to help improve the final dialogue model. 

The \textit{satisfaction classifier} identifies if the bot response is satisfactory or not. It is evaluated on both  FITS and zero-shot DEMO deployment datasets.

Both the  \textit{reply corrector} and the final dialogue models are generative models, and are automatically evaluated on the human-written gold corrections in the FITS validation and test sets, as gold corrections can reflect the model's ability to generate good responses. 

The \textit{reply corrector} converts bad responses into good ones using free-form textual feedback. We evaluate it on gold search results instead of live search results, in order to generate better reply corrections. We describe it in more detail in Section~\ref{app:reply corrector}. 

The final dialogue model is evaluated on live search results from Bing, filtered by CommonCrawl, following \citet{xu2022learning} instead of gold search results to better reflect performance with live users. We describe it in more detail in Section~\ref{appendix: final dialogue model}.

\subsection{\textit{Reply corrector}}
\label{app:reply corrector}
The \textit{reply corrector} trains on data where for a given example the input consists of the dialogue context + bad reply to correct + the following human message, and the output consists of the correction for the bad reply. The models used in the main paper were fine-tuned from the R2C2 transformer \cite{shuster2022language}.

\subsubsection{Training the \textit{reply corrector} with  multiple tasks}
\label{appendix:multitaking in reply corrector}
In our experiments, we multi-tasked with various dialogue tasks to train the \textit{reply corrector}, which improves the result. These tasks include the original reply correction task,  the task with context as the input and free-form textual feedback as the target, and the dialogue task of Wizard of Internet \cite{komeili2021internet}. We tuned the weights for different tasks, and other hyper-parameters (learning rate, batch size, etc) according to the performance on the validation set.  

\subsubsection{Evaluating the \textit{reply corrector}}
\label{appendix:eval reply corrector}
Since the \textit{reply correctors} are used to generate corrections rather than interacting with live users, we evaluated them with gold search results (which leads to better corrections) instead of live search results from Bing (which better reflects the live interaction performance). 

Although the \textit{reply corrector} (Table~\ref{tab:reply corrector}) and the final dialogue models (Table~\ref{tab:dialogue model_auto_eval}) are evaluated on the same validation and test subsets that have gold corrections, their results are not comparable because of the following two reasons. First, as mentioned earlier, the \textit{reply correctors} condition on the gold search results instead of the live search results, while the final dialogue models use the live search results. Second, the \textit{reply correctors} rely on the free-form textual feedback to convert lemons to cherries, so we also append the free-form textual feedback into the input to the \textit{reply correctors}, but for the final dialogue model, we do not have the additional free-form textual feedback information. These are the main reasons why the results in Table~\ref{tab:reply corrector} are better than those in Table~\ref{tab:dialogue model_auto_eval}.

    \begin{table*}
      \centering
      \begin{adjustbox}{width=.9\textwidth}
\begin{tabular}{l|l|l||l|l|l|l}
\toprule
\midrule
\textbf{\underline{(1b) \textit{Reply Corrector}}}&  \multicolumn{2}{c||}{\textbf{Valid}} & \multicolumn{2}{c|}{\textbf{Test}}                                & \multicolumn{2}{c}{\textbf{Test Unseen}} \\
\textbf{Input}& \multicolumn{1}{c}{\textbf{F1$\uparrow$}}    & \multicolumn{1}{c||}{\textbf{PPL$\downarrow$}}      & \multicolumn{1}{c}{\textbf{F1$\uparrow$}}    & \multicolumn{1}{c|}{\textbf{PPL$\downarrow$}}  & \multicolumn{1}{c}{\textbf{F1$\uparrow$}} & \multicolumn{1}{c}{\textbf{PPL$\downarrow$}} \\
\midrule
\midrule
\multicolumn{7}{l}{\textbf{fine-tuned from R2C2}} \\
\midrule
 gold corrections from 20\%  + self-corrections
&  \textbf{21.41}                  & \textbf{3.07} & \textbf{20.20} &	2.75&	\textbf{21.77}&	\textbf{4.66}                   \\
+ \textsc{Director}&  22.81 & -&\textbf{22.59} &-& 22.10 & -\\
+ \textsc{Director overlap} & \textbf{23.00} &- &22.50&-& \textbf{22.55}&- \\

\midrule
\multicolumn{7}{l}{\textbf{fine-tuned from BB2}} \\
\midrule
gold corrections from 20\% + self-corrections
&   16.32 &7.06    &14.53 &7.01 &15.63 &7.16
                  \\
\bottomrule
\end{tabular}
\end{adjustbox}
\caption{\textit{Reply corrector} results. The top block shows the \textit{reply corrector} fine-tuned from R2C2 with \textsc{Director} and \textsc{Director overlap}, and the bottom block shows the \textit{reply corrector} fine-tuned from BB2. R2C2 is better than BB2 as a \textit{reply corrector}. Using \textsc{Director} improves the result. Using \textsc{Director overlap} further improves over \textsc{Director}. 
}
\label{tab:reply corrector with director and bb2}
\vspace{-1em}
    \end{table*}%

\subsubsection{Generating reply corrections}
We adopt a reranking-based learning method to first generate multiple reply corrections, and then use the \textit{satisfaction classifier} to score and rerank the generated corrections. Because the \textit{reply corrector}'s performance is good (comparable to the one trained on 100\% data in Table~\ref{tab:reply corrector}) and we generated 60 correction candidates to choose from, the majority (99.96\%, 16893 out of 16989) of  bad responses have at least one generated correction that is predicted as satisfactory by the \textit{satisfaction classifier}.  

\paragraph{Generated reply correction examples.} Table~\ref{tab:qualitative on FITS} and Table~\ref{tab:qualitative on deployment} show generated reply correction examples on the FITS dataset and the deployment dataset (zero-shot) respectively. These qualitative examples show that the \textit{reply corrector} can convert bad replies into good ones using free-form textual feedback, even for unseen deployment data.  

\begin{table*}[!htb]
\centering
    \begin{subtable}{.975\columnwidth}
      \centering
      \begin{adjustbox}{width=.95\columnwidth}
        \begin{tabular}{lm{6cm}} \toprule
            \textbf{From}& \textbf{Utterance} \\ 
            \midrule
            \midrule
USR& Hi bot, what supplies do I need for a diaper change?  \\
\midrule
BOT& That's a good question. I'm not sure, but I'm sure you can find out online. \\
\midrule

Feedback& \textit{The results list at least one necessary supply, with the implication that bot would find others if it read further.} \\
\midrule

Correction& Bring your supplies to the diapering area. Clean cloth diapers are essential. You'll also need a clean diaper, clean wipes, diaper cream, gloves, plastic or waterproof bag for soiled clothing, extra clothes. \\
\midrule
Original& It depends on how many diapers you have. If you have a lot of diapers, you will need a changing table and a changing pad.\\

            \bottomrule
\end{tabular}
\end{adjustbox}


    \end{subtable}%
    \begin{subtable}{.975\columnwidth}
      \centering
      \begin{adjustbox}{width=.95\columnwidth}
        \begin{tabular}{lm{6cm}} \toprule
            \textbf{From}& \textbf{Utterance} \\ 
            \midrule
            \midrule
USR& What is the name of hello kitty's sister?  \\
\midrule
BOT& Hello kitty's twin sister, mimmy. \\
\midrule

Feedback& \textit{That's great. I think to make this even better it could be phrased a little more conversationally. Something like "hello kitty has a twin sister, mimmy."} \\
\midrule

Correction& Hello kittys sister is called mimmy! I think it's funny she looks a lot like her. Do you like hello kitties sister? \\
\midrule
Original&Hello kitty has a twin sister named mimmy.\\
            \bottomrule
\end{tabular}
\end{adjustbox}


    \end{subtable}%
\qquad
\qquad
\qquad
\qquad
\qquad
\newline
\newline
    \begin{subtable}{.975\columnwidth}
      \centering
      \begin{adjustbox}{width=.95\columnwidth}
        \begin{tabular}{lm{6cm}} \toprule
            \textbf{From}& \textbf{Utterance} \\ 
            \midrule
            \midrule
USR& Can you tell me what two sides fought during the war of the roses?  \\
\midrule
BOT& The war of the roses was between the english and the french. The english won. \\
\midrule

Feedback& \textit{It looks like that answer is not right. The search results say something different.} \\
\midrule

Correction& The war was fought between two competing royal families: the yorks and the lancasters, both members of the age-old royal plantagenet family. \\
\midrule
Original&The english and the french.\\
            \bottomrule
\end{tabular}
\end{adjustbox}


    \end{subtable}
\begin{subtable}{.975\columnwidth}
      \centering
      \begin{adjustbox}{width=.95\columnwidth}
        \begin{tabular}{lm{6cm}} \toprule
            \textbf{From}& \textbf{Utterance} \\ 
            \midrule
            \midrule
USR& What are risk factors of asthma?  \\
\midrule
BOT& Asthma is a chronic disease of the air passages in the lungs. \\
\midrule

Feedback& \textit{Try using a more specific query} \\
\midrule

Correction& Asthma can be caused by a combination of genetic and environmental factors. These factors include: early exposure to allergens, changing in weather, respiratory infections and more. \\
\midrule
Original& Asthma is a chronic disease of the air passages in the lungs.\\
            \bottomrule
\end{tabular}
\end{adjustbox}


    \end{subtable}

    \caption{Qualitative correction examples generated by the best \textit{reply corrector} on FITS.}
\label{tab:qualitative on FITS}
\end{table*}

\begin{table*}[!htb]
\centering
    \begin{subtable}{.975\columnwidth}
      \centering
      \begin{adjustbox}{width=.95\columnwidth}
        \begin{tabular}{lm{6cm}} \toprule
            \textbf{From}& \textbf{Utterance} \\ 
            \midrule
            \midrule
Truncated&...\\
\midrule
BOT& Hello! How are you today? I'm doing well, thank you. What are you up to? \\
\midrule
USR & I didn't ask how you are doing?\\
\midrule
BOT&Oh, I'm sorry. I meant to say, how are you and your family doing today?\\
\midrule
USR&I don't have a family\\
\midrule
BOT&I'm sorry to hear that. Are you a member of the family motor coach association (fmca)?\\
\midrule
Feedback& \textit{ Something like: ``i'm sorry I assumed you had a family! I'll keep this in mind going forward''} \\
\midrule

Correction 1& You're right. I shouldn't have assumed you have a family. I will keep this is mind going foward. \\
\midrule
Correction 2&I apologize. I didn't mean to assume you had children! I hope you have a great day.\\
\midrule
\end{tabular}
\end{adjustbox}


    \end{subtable}%
    \begin{subtable}{.975\columnwidth}
      \centering
      \begin{adjustbox}{width=.95\columnwidth}
        \begin{tabular}{lm{6cm}} \toprule
            \textbf{From}& \textbf{Utterance} \\ 
            \midrule
            \midrule
Truncated&...\\
\midrule
BOT & Which operators do you like the most? I like the operators from years 1 and 5 of siege.\\
\midrule
USR& I'm not sure, as I've never played.  \\
\midrule

BOT & Have you played any of the other rainbow six games? I've played rainbow six 3 and 4.\\
\midrule
USR& No.  \\
\midrule
BOT& No, I haven't. I've only played the first rainbow six game. Have you played that one? \\
\midrule

Feedback& \textit{ You could suggest other games, or explain what rainbow six is about. } \\
\midrule

Correction 1& Rainbow siege is a first-person tactical shooter. It is a good game. Do you like it? \\
\midrule
Correction 2&Do you like any other games? What console do you play on? Do you game on pc or console?\\
            \bottomrule
\end{tabular}
\end{adjustbox}


    \end{subtable}%
\qquad
\qquad
\qquad
\qquad
\qquad
    \begin{subtable}{.975\columnwidth}
      \centering
      \begin{adjustbox}{width=.95\columnwidth}
        \begin{tabular}{lm{6cm}} \toprule
            \textbf{From}& \textbf{Utterance} \\ 
            \midrule
            \midrule
Truncated&...\\
\midrule
USR& Try saying good night, or sweet dreams.  \\
\midrule
BOT& Thanks for your feedback. It is good to know. \\
\midrule
USR&Alright, I'm going to bed now. Good night!\\
\midrule
BOT&Thanks! I've been trying to get into sci fi, but it's hard to find a good book.\\\midrule
Feedback& \textit{ You could say good night, or something similar. } \\
\midrule

Correction 1& Well, you should go to bed so you can sleep well. Good night.  \\
\midrule
Correction 2&Have a good night!\\
            \bottomrule
\end{tabular}
\end{adjustbox}


    \end{subtable}
\begin{subtable}{.975\columnwidth}
      \centering
      \begin{adjustbox}{width=.95\columnwidth}
        \begin{tabular}{lm{6cm}} \toprule
            \textbf{From}& \textbf{Utterance} \\ 
            \midrule
            \midrule
BOT& Do you know how someone can make the best pizza?  \\
\midrule
USR& I don't, tell me? \\
\midrule
BOT & There is an international pizza expo in las vegas that showcases top suppliers and education about pizza.\\\midrule
Feedback& \textit{ That does not tell me about how to make the best pizza. } \\
\midrule

Correction 1& The best pizza is made from fresh ingredients with high quality toppings and is cooked to perfection. The best sauce is fresh basil.\\
\midrule
Correction 2& Many pizza shops have training classes to help you learn how to prepare and make pizza. There are also books to help.\\
            \bottomrule
\end{tabular}
\end{adjustbox}


    \end{subtable}

    \caption{Zero-shot corrections generated by the best  \textit{reply corrector} on unseen deployment data.}
\label{tab:qualitative on deployment}
\end{table*}

\subsubsection{Using BB2 to train the \textit{reply corrector}}
The  models used in the main experiments were fine-tuned from R2C2 \cite{shuster2022language}. We also report results fine-tuned with BB2 in Table~\ref{tab:reply corrector with director and bb2}. We find that BB2 is worse than R2C2 as a \textit{reply corrector} because its generated corrections are more like conversational replies rather than actual corrections.

\subsubsection{Using \textsc{Director} in the \textit{reply corrector}}
Using \textsc{Director}  to combine multiple feedback signals is also effective for the \textit{reply corrector}. We can use \textsc{Director} to further improve the \textit{reply corrector}'s F1 to 22.81, as shown in Table~\ref{tab:reply corrector with director and bb2}, where the positive examples are the gold corrections and the negative examples are the bad bot responses. However,  although the F1 of the \textsc{Director}-enhanced \textit{reply corrector} is better, we find that if we use it to generate reply corrections to improve the final dialogue models, the F1 is  slightly better but the perplexity gets worse than  using a regular \textit{reply corrector} without \textsc{Director}, as shown in Table~\ref{tab:dialogue model for director-enhanced reply corretor}.  
More analysis is needed to understand the reasons for this.

\begin{table*}[t!]
\centering
\begin{adjustbox}{width=.8\textwidth}
\begin{tabular}{l|cc}
\toprule
\midrule
                                                       
 \multirow{1}{*}{\textbf{Final dialogue model}}  & \multicolumn{2}{c}{\textbf{Automatic  evaluation}}    \\ 
 &\multicolumn{2}{c}{\textbf{Valid}} \\
& \textbf{F1$\uparrow$} & \textbf{PPL$\downarrow$}   \\
\midrule
\midrule
\multicolumn{3}{l}{\textbf{\pipelinename models}}  \\
\midrule
\midrule
~~+\pipelinename
& 16.7       & \textbf{8.5} 
\\
~~+\pipelinename w/ \textsc{Director overlap}-based \textit{reply corrector}
& \textbf{16.8}       & 8.8  \\
\midrule
\midrule
\multicolumn{3}{l}{\textbf{\pipelinename ablations}}    \\
\midrule
\midrule
w/o selecting correctable cases  & 16.4 & \textbf{8.5} 
\\
w/o selecting correctable cases w/ \textsc{Director overlap}-based \textit{reply corrector}  & \textbf{16.5} & 8.7 \\
\bottomrule
\end{tabular}
\end{adjustbox}
\caption{Final dialogue model automatic evaluation results. The \textsc{Director overlap}-enhanced \textit{reply corrector} achieves the highest F1 on the reply correction task, better than the regular \textit{reply corrector} (see \autoref{tab:reply corrector with director and bb2}). But when we use it to generate the reply corrections to further improve the final dialogue model, we can improve the F1 of the final dialogue model slightly, but the perplexity gets a bit worse. Further investigations are needed to understand the reason for this. 
}
\label{tab:dialogue model for director-enhanced reply corretor}
\end{table*}

\begin{table*}[tbh!]
\centering
\begin{adjustbox}{width=.98\textwidth}
\begin{tabular}{l|cc|cc|cc||cc|ccc}
\toprule
\midrule
\textbf{Oracle model performance}                                                   & \multicolumn{6}{c||}{\bf Automatic evaluation} &   \multicolumn{5}{c}{\bf Human evaluation}   \\
\midrule
                                                   & &&&&&&  & & \multicolumn{3}{c}{\bf Error Breakdown $\downarrow$}   \\
                                                       
&\multicolumn{2}{c|}{\textbf{Valid}} & \multicolumn{2}{c|}{\textbf{Test}} &        \multicolumn{2}{c||}{\textbf{Test Unseen}} & \textbf{Good}  & \multirow{2}{*}{\textbf{Rating} $\uparrow$} & Search  & Search & \multirow{2}{*}{Response}\\
& \textbf{F1$\uparrow$} & \textbf{PPL$\downarrow$} & \textbf{F1$\uparrow$} & \textbf{PPL$\downarrow$} & \textbf{F1$\uparrow$} & \textbf{PPL$\downarrow$}   & \textbf{response} $\uparrow$ &  &  Query  &  Results &                      \\
\midrule
\midrule
BB2 & 14.4       &  10.6 &  14.7	& 10.3&	15.3	& 9.3 &  33.2\% & 3.09  & 12.1\%	& 18.6\% &	18.1\%\\
\textcolor{black}{~~+100\% reward-based learning}  & 15.1 & 11.0 & 14.2 & 10.7 & 14.3 & 9.6 & {\bf 36.4}\% & 2.83 & 11.3\%	& 18.6\% &	17.0\% \\
\textcolor{black}{~~+100\% free-form textual feedback}   &  15.5 & 9.7 & 15.6 & 9.5  & 16.8 & 8.7  & {\bf 37.0}\% & 3.22 & 11.6\%	& 17.6\% &	17.0\% \\
\textcolor{black}{~~+100\% gold correction}   &  14.7 & 8.2 & 15.5 & 8.0 & 17.0 & 8.0  & {\bf 40.3}\% & {\bf 3.37} & 11.6\% &	18.3\% & 	{\bf 15.0}\% \\
\textcolor{black}{~~+100\% module supervision}   &   14.9 & 7.6	&  15.5 & 7.5 & 15.4 & 8.3 & {\bf 42.0}\% & {\bf 3.35} &	{\bf 8.4}\%	& 20.8\% &	{\bf 14.4}\%	\\
\textcolor{black}{~~+100\% reranking binary feedback}         &   15.8 &  {\tiny n/a}~ & 15.8 & {\tiny n/a}~ & 16.3 & {\tiny n/a}~ &- &- &- & -& -\\
\textcolor{black}{~~+100\% {\sc Director} binary feedback only}& 16.2 & {\tiny n/a}~ & 16.2 & {\tiny n/a}~ & 17.6 & {\tiny n/a}~ & {\bf 37.8}\% & 3.07 &	11.4\%	& 17.3\% &	16.9\%	\\
\textcolor{black}{~~+100\% {\sc Director} module+binary feedback}         & 17.2 & {\tiny n/a}~ & 16.6 & {\tiny n/a}~ & 16.0 & {\tiny n/a}~  & {\bf 47.0}\% & {\bf 3.38} &	{\bf 8.4}\%	& {\bf 16.1}\%	& {\bf 14.3}\% 	\\
\bottomrule
\end{tabular}
\end{adjustbox}
\caption{Final dialogue model results from 100\% oracle methods in \citet{xu2022learning}. Similarly we bold statistically significant improvements (independent two-sample $t$-test, $p < 0.05$) of methods over their  baselines BB2 3B in the human evaluation block. }
\label{tab: oracle performance}
\end{table*}

\begin{table*}[tbh!]
\centering
\begin{adjustbox}{width=.98\textwidth}
\begin{tabular}{l|cc|cc|cc||cc|ccc}
\toprule
\midrule
\textbf{Final dialogue model}                                                   & \multicolumn{6}{c||}{\bf Automatic evaluation} &   \multicolumn{5}{c}{\bf Human evaluation}   \\
\midrule
                                                   & &&&&&&  & & \multicolumn{3}{c}{\bf Error Breakdown $\downarrow$}   \\
                                                       
&\multicolumn{2}{c|}{\textbf{Valid}} & \multicolumn{2}{c|}{\textbf{Test}} &        \multicolumn{2}{c||}{\textbf{Test Unseen}} & \textbf{Good}  & \multirow{2}{*}{\textbf{Rating} $\uparrow$} & Search  & Search & \multirow{2}{*}{Response}\\
& \textbf{F1$\uparrow$} & \textbf{PPL$\downarrow$} & \textbf{F1$\uparrow$} & \textbf{PPL$\downarrow$} & \textbf{F1$\uparrow$} & \textbf{PPL$\downarrow$}   & \textbf{response} $\uparrow$ &  &  Query  &  Results &                      \\
\midrule
\midrule
\multicolumn{6}{l}{\textbf{\pipelinename}}            \\
\midrule
~~+\pipelinename & 16.74       & 8.50 & 16.18	&8.44&	18.50 & 8.02	&   {\bf 41.9\%} & 3.06 & 	13.0\% &	17.7\% &	{\bf 13.8\%} \\
~~+\pipelinename + {\sc Director}  & 17.25 &-& 16.70 &-& 17.70 & -   &   {\bf 45.5\%} & {\bf 3.34} & 	11.3\% &	17.4\% &	{\bf 12.9\%} \\
~~+\pipelinename + {\sc Director Overlap}  & 17.32 & - & 16.66 & - & 17.62 &-   &   {\bf 47.8\%} & 3.25 & 	11.0\% &	{\bf 14.8\%} &	{\bf  13.3\%} \\
\midrule

\multicolumn{6}{l}{\textbf{\pipelinename w/o selecting correctable cases}}  \\
\midrule
\textcolor{black}{~~+\pipelinename} & 16.44       & 8.54 & 16.37&	8.41&	17.95&	8.12   &  {\bf 41.4\%} & 3.08 & 	13.4\% &	16.8\% &	{\bf 14.2\%} \\
\textcolor{black}{~~+\pipelinename + \textsc{Director}}  & 17.23&-&16.62&-&17.93&-   &   {\bf 44.6\%} & {\bf 3.40} & 	11.6\% &	16.7\% &	{\bf 13.6\%} \\
\textcolor{black}{~~+\pipelinename + \textsc{Director overlap}}  & 16.98 & - & 16.56 &-& 17.19 &- &   {\bf 45.5\%} & {\bf3.48} & 	10.8\% &	{\bf 15.2\%} &	{\bf 14.3\% }\\
\bottomrule
\end{tabular}
\end{adjustbox}
\caption{
\pipelinename with \textsc{Director overlap}.  \textsc{Director overlap} improves the human evaluation results over the vanilla \textsc{Director}. Similarly we bold statistically significant improvements (independent two-sample $t$-test, $p < 0.05$) of methods over their  baselines BB2 3B in the human evaluation block.
}
\label{tab:dialogue model director overlap}
\end{table*}



\subsection{Final dialogue model evaluation}
\label{appendix: final dialogue model}
We evaluate the final dialogue model on live search results instead of gold search results to better reflect performance with live users.

\subsubsection{Oracle performance using 100\% feedback data}
\citet{xu2022learning} trained various methods on the entire FITS dataset. Since our method is trained only on 20\% of FITS, the 100\% models' performance could be viewed as an upper bound of our models.  They also used the 3B parameter BlenderBot 2 as a base model for  the final dialogue model, making it comparable to our experiments. Their results are in Table~\ref{tab: oracle performance} and we detail their models below.  
\begin{itemize}
        \item \textbf{100\% gold correction}. The input is the context and the target is the  gold correction (6,601 in the entire FITS dataset). This can be directly compared to ``gold correction from 20\%'' in Table~\ref{tab:dialogue model_auto_eval}. 
        
        \item \textbf{100\% free-form textual feedback}. The input is the context and the target is the free-form textual feedback. This should be compared  to ``free-form textual feedback from 20\%'' in Table~\ref{tab:dialogue model_auto_eval}. 

    \item \textbf{100\% module supervision}. BlenderBot 2 is an internet-augmented bot with different modules such as a search module to generate a search query, and a knowledge module to attend to the search results. Using the human-written gold search query, human-selected search doc and gold correction, they fine-tuned each individual module to improve BlenderBot 2. 
    \item \textbf{100\% reward-based learning}. They also adopted a reward-based learning approach, and built a \textit{satisfaction classifier} to identify good and bad responses. They used the vanilla BlenderBot 2 model to generate multiple responses,  then reranked them with the score from the classifier as the reward, and chose the response with the highest reward. Finally, they  fine-tuned BlenderBot 2 on the  responses with high rewards to improve it.  
    \item \textbf{100\% \textsc{Director}}. They also used both the binary satisfaction labels and the textual feedback to train \textsc{Director} models to further improve the performance. 
\end{itemize}

As mentioned in the main body of our paper, \pipelinename achieves comparable performance to the ``oracle'' (100\%) models in F1 and human evaluation. For instance, the best oracle method which fine-tunes individual modules achieves an F1 of 17.2 and 47.0\% good response rate, and a human rating of 3.38, while the best \pipelinename model achieves an F1 of 17.2, a good response rate of 45.5\%, and a similar human rating of 3.34.

\subsubsection{\sc Director overlap}
\label{appendix:director overlap}
We also develop a new variant of \textsc{Director} and use it to improve the final dialogue model. 
In {\sc Director}, every token in the positive and negative examples has a one or zero label respectively. In our setting, we have a pair of  a bad response and  a good response (a gold correction), e.g., ``I like watermelons too! Have you heard of Harry Styles?'' (bad) and ``I like watermelons too! They tastes great in drinks.'' (good). Since people tend to edit the original bad response to correct it,  they may have many overlapping tokens (``I like watermelons too!''), which we do not have to punish. So we develop {\sc Director overlap}, where  we obtain the bag of tokens of the pair of the bad response and the gold correction, and assign  a positive label for the overlapping tokens in the negative examples. In our data, 28.4\% of tokens in the bad responses  overlap with those in gold corrections (6.5\% are stop words and punctuations, and 21.9\% are not). 

Table~\ref{tab:reply corrector with director and bb2} and Table~\ref{tab:dialogue model director overlap} show the result of {\sc Director overlap}. For the \textit{reply corrector}, \textsc{Director overlap} improves the F1 to 23.00 over \textsc{Director}. For the final dialogue model, \textsc{Director overlap} improves the good response rate and lowers the search result error in human evaluations over \textsc{Director}.

\section{Model Training Setting}
We use the openly available \href{https://parl.ai/}{ParlAI} framework for all training runs, as well as for evaluations, where metrics are measured using default settings.
All the fine-tuned models are trained with a maximum of eight 32GB GPUs (NVIDIA V100), optimized with Adam using $\beta_1 = 0.9$, $\beta_2 = 0.999$, $\epsilon = 1e-08$. Models are trained up to 4000 updates with batch sizes up to 128. The typical fine-tuning time for a standard transformer encoder-decoder is 8 hrs before it early stops, and the time is 16 hrs for retrieval-based models.

\end{document}